%% file: arxiv.tex
\documentclass{article}
\usepackage{amsmath}

\usepackage[preprint]{neurips_2025}

\usepackage[utf8]{inputenc} 
\usepackage[T1]{fontenc}    
\usepackage{hyperref}       
\usepackage{url}            
\usepackage{booktabs}       
\usepackage{amsfonts}       
\usepackage{nicefrac}       
\usepackage{microtype}      
\usepackage{multirow}
\usepackage[table]{xcolor}
\definecolor{brighterblue}{RGB}{0, 50, 150} 
\definecolor{myblue}{HTML}{377DBD}
\usepackage{graphicx}
\usepackage{float}
\usepackage{enumitem}
\usepackage{cleveref}

\usepackage{makecell}
\usepackage{tikz}
\usepackage{wrapfig}
\usepackage{pifont}

\title{UniVG-R1: Reasoning Guided Universal Visual Grounding with Reinforcement Learning}

\author{Sule Bai$^{1,2,*}$, Mingxing Li$^2$, Yong Liu$^1$, Jing Tang$^2$, Haoji Zhang$^1$,\\ \textbf{Lei Sun$^2$\textsuperscript{$\ddagger$}, Xiangxiang Chu$^2$, Yansong Tang$^1$\textsuperscript{$\dagger$}} \\ $^1$Tsinghua Shenzhen International Graduate School, Tsinghua University\\ $^2$AMAP, Alibaba Group\\ 
{\tt\small \{bsl23@mails.,tang.yansong@sz.\}tsinghua.edu.cn}
}

\begin{document}

\def\thefootnote{}\footnotetext{
* Work done during the internship at AMAP, Alibaba Group. $\quad$ $\dagger$ Corresponding author $\quad$ $\ddagger$ Project lead
}

\maketitle

\begin{abstract}
Traditional visual grounding methods primarily focus on single-image scenarios with simple textual references. However, extending these methods to real-world scenarios that involve implicit and complex instructions, particularly in conjunction with multiple images, poses significant challenges, which is mainly due to the lack of advanced reasoning ability across diverse multi-modal contexts.
In this work, we aim to address the more practical universal grounding task, and propose UniVG-R1, a reasoning guided multimodal large language model (MLLM) for universal visual grounding, which enhances reasoning capabilities through reinforcement learning (RL) combined with cold-start data.
Specifically, we first construct a high-quality Chain-of-Thought (CoT) grounding dataset, annotated with detailed reasoning chains, to guide the model towards correct reasoning paths via supervised fine-tuning.  Subsequently, we perform rule-based reinforcement learning to encourage the model to identify correct reasoning chains, thereby incentivizing its reasoning capabilities. 
In addition, we identify a difficulty bias arising from the prevalence of easy samples as RL training progresses, and we propose a difficulty-aware weight adjustment strategy to further strengthen the performance. 
Experimental results demonstrate the effectiveness of UniVG-R1, which achieves state-of-the-art performance on MIG-Bench with a 9.1\% improvement over the previous method. Furthermore, our model exhibits strong generalizability, achieving an average improvement of 23.4\% in zero-shot performance across four image and video reasoning grounding benchmarks. The project page can be accessed \href{https://amap-ml.github.io/UniVG-R1-page/}{\textcolor{myblue}{here}}.
\end{abstract}

\section{Introduction}

\begin{figure}
    \centering
    \includegraphics[width=\linewidth]{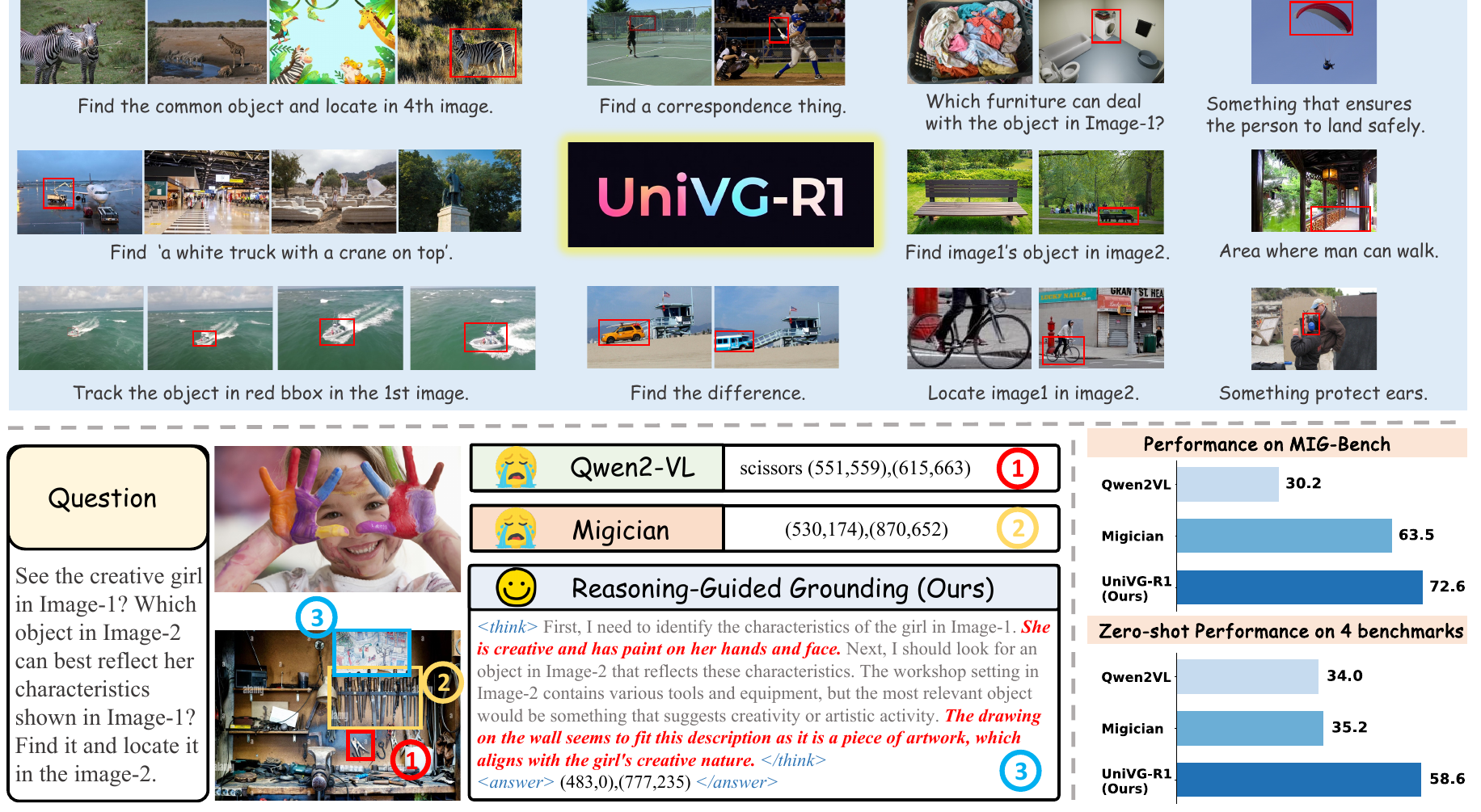}
    \vspace{-10pt}
    \caption{\textbf{UniVG-R1} tackles a wide range of visual grounding tasks with complex and implicit instructions. By combining GRPO training with a cold-start initialization, it effectively reasons over instructions and visual inputs, significantly improving grounding performance. Our model achieves state-of-the-art results on MIG-Bench and exhibits superior zero-shot performance on four reasoning-guided grounding benchmarks with an average 23.4\% improvement.}
    \label{fig:teaser}
    \vspace{-3pt}
\end{figure}

Visual grounding is a significant task that aims to recognize and localize target regions in images with the guidance of instructions.
Conventional setting~\cite{yu2016refcoco, mao2016refcoco} typically localizes objects based on predefined categories or explicit simple instructions (e.g., ``the blue shirt'').
It struggles to perform comprehension of implicit user instructions jointly with complex visual contexts. For example, handling nuanced queries like ``Which furniture in Image-2 can deal with the objects in Image-1?'' (as shown in \Cref{fig:teaser}) requires advanced reasoning of user instructions across multiple images. 
Therefore, we focus on achieving universal visual grounding by unlocking a broader spectrum of
challenging scenarios in this work.

To effectively tackle this universal and sophisticated visual grounding task, the ability to reason complex and implicit  correspondence across diverse visual contexts is crucial. 
However, most previous works~\cite{yu2016refcoco, mao2016refcoco,xiao2024hivg,deng2021transvg} have focused on localizing targets within single-image scenarios with intuitive instructions, which demonstrates a remarkable divergence from the requirements commonly observed in real-world applications.
With the development of Large Language Models (LLMs), some works~\cite{lai2024lisa,wang2024llmseg,chen2023shikra,zhan2024griffon} propose to leverage the powerful comprehension ability of LLMs to facilitate grounding task.
Despite the great progress in understanding text instructions, these works are still limited to single image scenarios and fail to incorporate modeling of correlations across multiple images. 
Recently, 
Migician~\cite{li2025migician} introduces a multi-image grounding benchmark encompassing diverse grounding tasks, thereby advancing foundational initiatives to bridge this research gap.
However, Migician does not incorporate an explicit reasoning process during training, thereby falling short in terms of \textit{advanced reasoning capabilities}, particularly in handling \textit{complex and implicit instructions across diverse images} that are essential for universal visual grounding.



Recognizing these limitations, we draw inspirations from the recent success of large reasoning models~\cite{jaech2024openaio1, guo2025deepseek, team2025kimi}, such as DeepSeek-R1~\cite{guo2025deepseek}, which employs rule-based reinforcement learning (RL) to significantly enhance large language model performance in solving challenging problems requiring in-depth reasoning. 
To this end, we explore the potential of the RL paradigm in this work and present UniVG-R1, a powerful reasoning guided MLLM designed for universal grounding. Specifically, we initially conduct experiments using pure RL on recent advanced MLLMs (e.g., Qwen2-VL~\cite{wang2024qwen2vl}), but find that it struggles to generate correct reasoning, leading to suboptimal performance. 
We ascribe this limitation to inherent constraints in the model's intrinsic knowledge base when handling multi-image contexts, which critically hinders effective exploration of the reasoning space solely through RL.
To address this limitation, we construct a high-quality Chain-of-Thought (CoT) \cite{wei2022chain} grounding dataset comprising 90k samples across diverse tasks, each annotated with reasoning chains and further validated by MLLMs to ensure correctness. Based on this dataset, we employ a two-stage training protocol. The first stage utilizes a cold-start supervised fine-tuning training, which directs the model towards correct reasoning pathways, then it is followed by a Group Relative Policy Optimization (GRPO) training with an IoU-based verifiable reward functions, further incentivizing the model's reasoning capabilities.

Furthermore, we identify an inherent difficulty bias in the GRPO algorithm. Since GRPO computes the relative advantage within each group by normalization, it overlooks the varying difficulty among different samples. Consequently, easier samples receive policy gradient updates of a magnitude similar to that of more challenging, lower-performing samples. This bias diminishes the training efficiency, especially as the proportion of easy samples increases during the RL training. To address this issue, we propose a simple online difficulty-aware weight adjustment strategy that dynamically scales the gradients of samples based on their difficulty, thereby encouraging more policy gradient updates from harder samples. We experiment with multiple difficulty metrics and empirically find that all variants consistently yield additional performance improvements.



With the above designs modeling and consolidating reasoning abilities for diverse correspondence, our UniVG-R1 is capable of effectively addressing complex multimodal contexts, facilitating versatile and generalizable visual grounding applications in real-word scenarios. To demonstrate the effectiveness of our method, we conduct extensive evaluations on MIG-Bench~\cite{li2025migician}, achieving state-of-the-art results with an average improvement of more than 9\% on ten tasks.
Furthermore, our model demonstrates superior generalizability, evidenced by significant zero-shot performance gains on a group of benchmarks: +27.8\% on LISA-Grounding~\cite{lai2024lisa}, +15.9\% on LLMSeg-Grounding~\cite{wang2024llmseg}, +20.3\% on ReVOS-Grounding~\cite{yan2024visa}, and  +25.3\% on ReasonVOS~\cite{bai2024videolisa}.

In summary, we make the following contributions: (1) We propose UniVG-R1, a reasoning guided MLLM for universal visual grounding, which employs GRPO training combined with a cold-start initialization to effectively enhance reasoning capabilities across multimodal contexts. 
(2) A high-quality CoT grounding dataset is introduced, encompassing diverse tasks, each meticulously annotated with detailed reasoning chains to facilitate advanced reasoning-based grounding.
(3) We identify a difficulty bias in GRPO training, and propose a difficulty-aware weight adjustment strategy.
Experiments validate that GRPO equipped with this strategy consistently enhance the model performance. (4) Extensive experiments demonstrate that our model achieves state-of-the-art performance across multiple grounding benchmarks, showcasing its versatility and generalizability.

\section{Related Work}
\subsection{Visual Grounding}
Visual grounding involves localizing a visual element in an image based on a specific linguistic query, which has broad applications across many tasks~\cite{yang2024language,liu2024universal,liu2024open,bai2024self,ye2024voco,ye2024atp,zhang2024narrative,liu2024coarse,wang2024ponder,zhang2024flash}. RefCOCO/+/g~\cite{yu2016refcoco, mao2016refcoco, kazemzadeh2014referitgame, qiao2020referring} is a widely used benchmark for this task. Given an image and a referring expression (e.g., “the left apple”), the model is required to identify the referred object. Early approaches~\cite{xiao2023clipvg, xiao2024hivg, liu2024groundingdino, yan2023uninext,deng2021transvg,li2022glip,jin2023refclip,subramanian2022reclip} leverage vision-language pre-trained models such as CLIP~\cite{radford2021learning} to improve fine-grained understanding. With the rapid development of multimodal large language models (MLLMs)~\cite{liu2023llava, wang2024qwen2vl, li2024llavaov, chu2024mobilevlm}, researchers have introduced more challenging datasets~\cite{lai2024lisa, bai2024videolisa, wang2024llmseg, yan2024visa}, such as LISA-Grounding~\cite{lai2024lisa}, which require models to comprehend complex instructions (e.g., “find the food rich in vitamins in the image”). A series of works~\cite{wei2025lenna, chen2023shikra, li2024groundinggpt, zhan2024griffon, you2023ferret} have been proposed to address these tasks. Recently, Migician~\cite{li2025migician} introduces a free-form multi-image grounding task, which requires models to perform multi-context understanding and grounding across ten different subtasks, including static difference, common object, and correspondence.
However, existing methods lack advanced reasoning capabilities, resulting in suboptimal performance when dealing with complex multimodal contexts. In this work, we aim to enhance the model’s reasoning ability by introducing reasoning chains, thereby improving its performance in challenging scenarios.

\subsection{Reasoning-Chain Guided Reinforcement Learning}
Reinforcement Learning (RL) has emerged as a pivotal research direction for enhancing the complex reasoning capabilities of Large Language Models (LLMs)~\cite{guo2025deepseek, shao2024deepseekmath, jaech2024openaio1, team2025kimi, chu2025gpg, reft,yang2024qwen2, hui2024qwen25coder,jiao2024preference,zhang2024o1,ying2024internlmmath}. OpenAI-o1~\cite{jaech2024openaio1} employs Reinforcement Learning from Human Feedback (RLHF) during fine-tuning, which significantly enhances the model’s reasoning ability and alignment with human preferences. The recent DeepSeek-R1~\cite{guo2025deepseek} employs GRPO~\cite{shao2024deepseekmath}, which, unlike traditional RL algorithms that rely on a critic model, directly utilizes rule-based verifiable rewards to guide the model's reasoning process. This approach significantly simplifies the training procedure and has proven highly effective in eliciting reasoning capabilities. Group policy gradient \cite{chu2025gpg} (GPG) further simplifies the pipeline and performs better.
This trend is gradually extending to MLLMs to further enhance their visual reasoning abilities~\cite{xu2024llavacot, liu2025segzero, huang2025visionr1, meng2025mmeureka, yu2025perceptionr1, yang2025r1onevision, ma2025deepperception,chen2025finger,peng2025lmmr1,deng2025boosting,zhou2025r1,deng2025openvlthinker,thinkbot}. Studies such as Visual-RFT~\cite{Visual-rft} and VLM-R1~\cite{vlm-r1} show that, for single-image visual grounding tasks, directly applying GRPO with a small number of samples can achieve improvements that surpass those of supervised fine-tuning. Vision-R1~\cite{huang2025visionr1} demonstrates the effectiveness of this approach in multimodal math benchmarks. In this work, we aim to extend this paradigm to the aforementioned universal grounding task.

\section{Method}
\begin{figure}
    \centering
    \includegraphics[width=\linewidth]{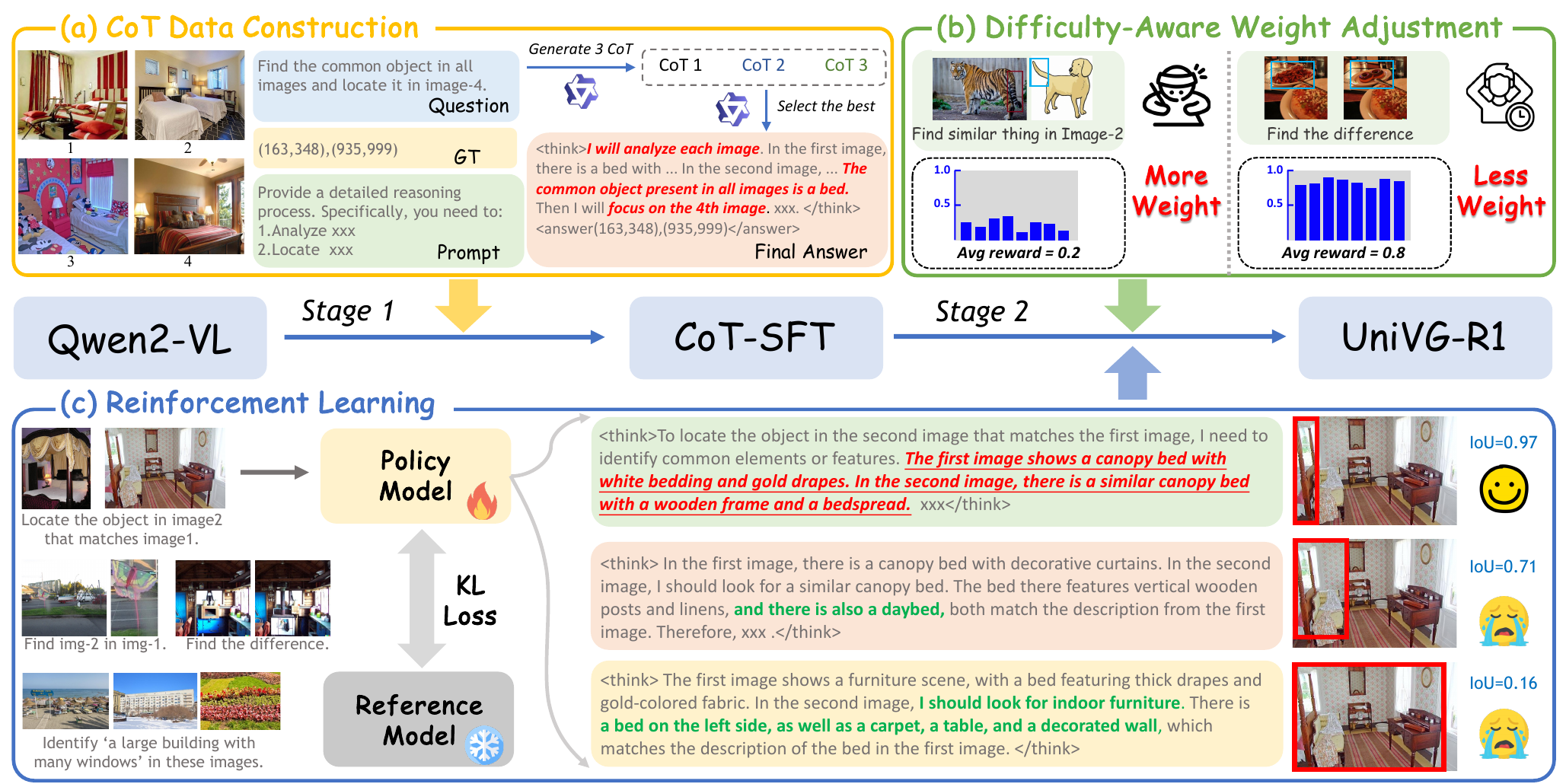}
    \vspace{-10pt}
    \caption{We adopt a two-stage training process. The first stage employs CoT-SFT, with the training data construction shown in (a). The second stage utilizes GRPO equipped with a difficulty-aware weight adjustment strategy in (b). The GRPO training process is illustrated in (c), where the policy model generates multiple responses, and each is assigned a distinct reward.}
    \label{fig:method}
\end{figure}

\subsection{Overview} 
In this section, we provide an overview of our proposed method, UniVG-R1. The task we address is a practical and universal visual grounding problem, where the model is tasked with localizing objects based on implicit and complex instructions within a multi-image context. Formally, given a textual instruction $T$, a target image $I$, and several additional images $V$, the model $\mathcal{M}$ is expected to output a bounding box $B$, defined as $B = \mathcal{M}(T, I, V)$.

Previous visual grounding methods typically rely on bounding box coordinate annotations or simple factual descriptions. After supervised fine-tuning, these models are restricted to such coordinates and lack explicit reasoning processes. However, the universal visual grounding task we address necessitates the model to comprehend complex instructions and additional visual inputs to perform localization. Motivated by the recent advancements in large reasoning models~\cite{jaech2024openaio1,guo2025deepseek,team2025kimi}, we aim to introduce this paradigm into our approach.

Our training process consists of two stages as shown in \Cref{fig:method}. In the first stage, we construct a high-quality dataset with Chain-of-Thought (CoT) annotations for supervised fine-tuning (SFT), enabling the model to learn structured reasoning trajectories. In the second stage, we employ rule-based reinforcement learning GRPO to guide the model in selecting correct reasoning chains, thereby further enhancing its reasoning capabilities. Additionally, we introduce a difficulty-aware weight adjustment strategy to enhance the model's performance during the GRPO training.

\subsection{Cold Start Data Construction and Chain-of-Thought Supervised Fine-tuning}\label{sec:cold-start}
Inspired by DeepSeek-R1-Zero~\cite{guo2025deepseek}, we initially explore the feasibility of training the model using pure reinforcement learning. However, experimental results in \Cref{sec:ablation} show that under the same amount of data, the model's performance is inferior to that achieved by supervised fine-tuning. We attribute this to the model's limited grounding ability in multi-image scenarios, which makes it challenging to explore the reasoning space solely through reinforcement learning. Therefore, it is necessary to construct a high-quality cold-start dataset in advance to guide the model's learning and endow it with grounding-oriented cognitive capabilities.


To this end, we randomly sample items from the MGrounding-630k dataset~\cite{li2025migician} and utilize the advanced multimodal large language model Qwen-VL-MAX~\cite{wang2024qwen2vl} to generate chain-of-thought reasoning processes. Specifically, as illustrated in \Cref{fig:method} (a), we provide the model with the question, bounding box coordinates, and a predefined CoT prompt, prompting it to generate reasoning processes in the format: \textit{``<think>thinking process here</think><answer>(x1, y1), (x2, y2)</answer>''}. For each item, we generate three reasoning chains and then use Qwen-VL-MAX to evaluate and select the best one as the final answer. Ultimately, we collect 76k samples and conduct further manual verification by randomly sampling 10\% of the data for human evaluation, achieving a final acceptance rate of 99.87\%. Please refer to the supplementary materials for more details. 

This dataset is subsequently utilized for supervised fine-tuning on Qwen2-VL-7B, resulting in our stage-1 model. The trained model is capable of producing final bounding box predictions through a coherent, step-by-step reasoning process.

\subsection{Reinforcement Learning for Enhancing Reasoning Capability} 
In the second stage, we employ rule-based reinforcement learning to enhance the model's reasoning abilities. Specifically, we adopt the Group Relative Policy Optimization (GRPO) algorithm~\cite{shao2024deepseekmath}. Unlike previous methods~\cite{ppo} that rely on an additional critic model, GRPO leverages a direct verification function to assess the correctness of each answer. Given a question $q$, the GRPO algorithm samples $N$ responses $\{o_1, o_2, \ldots, o_N\}$ from the policy model $\pi_{\theta_{old}}$, and evaluates each response using a rule-based verifiable reward function $R(q, o_i)$. 
For our task, we utilize two reward functions as described below:

\textbf{Accuracy Reward} ($r^{acc}$): Given the ground truth bounding box coordinates \( B_{GT} \) and the model's predicted coordinates denoted as \( B_{pred} \), we define the accuracy reward as $\mathbf{IoU}(B_{pred}, B_{GT})$, where $\mathbf{IoU}$ denotes the Intersection over Union metric. This reward encourages the model to generate bounding boxes that closely match the ground truth.

\textbf{Format Reward} ($r^{format}$): This reward ensures that the model's response strictly adheres to the required format. Specifically, the model must output: \textit{``<think>thinking process here</think><answer>(x1, y1), (x2, y2)</answer>''}, and this reward returns a value of 1 if the format is correct and 0 otherwise.

The total reward for a response $o_i$ is defined as $r_i = r^{acc}_i + r^{format}_i$. To determine the relative quality of these responses, GRPO normalizes the rewards by computing their mean and standard deviation.  The advantage for each response is then computed as: 
\begin{equation}
    A_i = \frac{r_i - \text{mean}(\{r_1, r_2, \dots, r_G\})}{\text{std}(\{r_1, r_2, \dots, r_G\})}.
\end{equation}

where $A_i$ represents the advantage of the candidate response $o_i$ relative to the other sampled responses within the group. GRPO encourages the model to generate responses with higher advantages by updating the policy $\pi_\theta$ to maximize the following objective function:
\begin{equation}
        \mathcal{J}_{GRPO}(\theta) = \mathbb{E}_{q \sim P(Q), \{o_i\}_{i=1}^N \sim \pi_{\theta_{old}}(O|q)} \\
    \frac{1}{N}\sum_{i=1}^N \frac{\pi_{\theta}(o_i|q)}{\pi_{\theta_{old}}(o_i|q)}A_i - \beta\mathbb{D}_{KL}(\pi_{\theta}||\pi_{ref})
\end{equation}

\begin{equation}
    \mathbb{D}_{K L}\left(\pi_{\theta} \| \pi_{r e f}\right)=\frac{\pi_{r e f}\left(o_{i} | q\right)}{\pi_{\theta}\left(o_{i} | q\right)}-\log \frac{\pi_{r e f}\left(o_{i} | q\right)}{\pi_{\theta}\left(o_{i} | q\right)}-1
\end{equation}

where $\beta$ is a hyperparameter that controls the degree of the KL loss. During the second stage of training, we add the prompt \textit{``First output the thinking process in <think> </think> tags and then output the bounding box in <answer> </answer> tags.''} to each question. The GRPO algorithm guides the model to select the correct reasoning chain from multiple sampled responses by assigning distinct advantages, thereby enhancing its reasoning capabilities, as shown in \Cref{fig:method} (c).

\subsection{Difficulty-Aware Weight Adjustment Strategy}\label{sec:difficulty}

During the stage 2 reinforcement learning process, we observe that most samples progressively become easier for the model, with the proportion of easy samples increasing and the proportion of 
\begin{wrapfigure}{r}{6cm}  
    \centering
    \includegraphics[width=6cm]{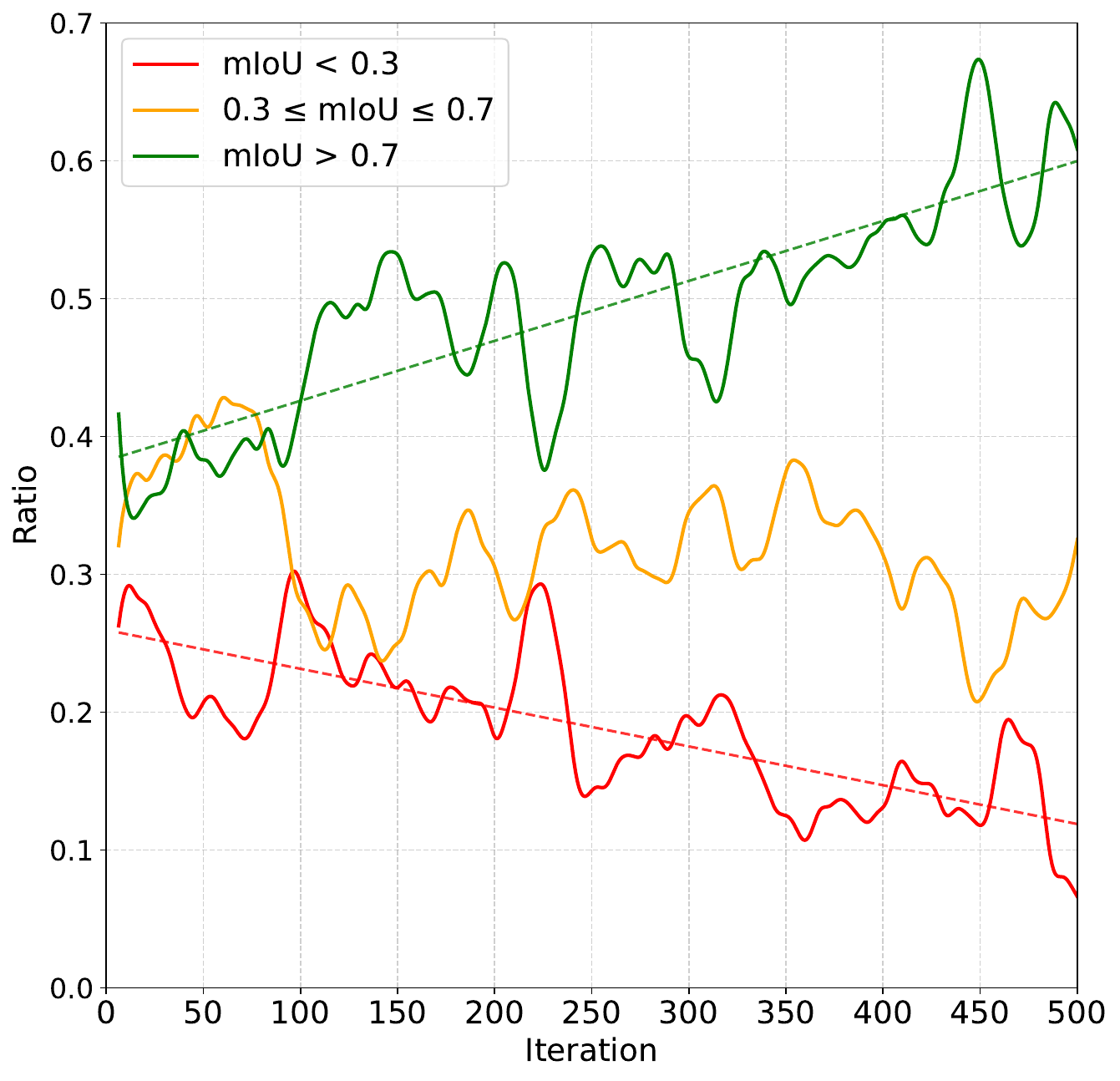}
    \vspace{-20pt}
    \caption{The proportion of easy, medium, and hard samples during GRPO training.}
    \label{fig:training_step}
    \vspace{-10pt}
\end{wrapfigure}
hard samples steadily decreases. If we define $\mathit{mIoU} = \mathtt{mean} (r_1^{acc}, r_2^{acc}, \ldots, r_G^{acc})$, where $\mathit{mIoU}$ is the average accuracy reward of all responses for a given sample. As shown in \Cref{fig:training_step}, the proportion of easy samples ($\textit{mIoU} > 0.7$) gradually increases, while the proportions of medium-difficulty samples ($0.3 < \textit{mIoU} < 0.7$) and hard samples ($\textit{mIoU} < 0.3$) both exhibit a declining trend. Since the GRPO algorithm normalizes rewards to calculate the relative advantage within each group, easy samples (e.g., $\textit{mIoU} = 0.8$) receives the same policy gradient update as hard samples (e.g., $\textit{mIoU} = 0.2$). This leads to a difficulty-bias issue. In particular, during the later stages of training, as easy samples become predominant, most updates are derived from these easier instances, making it difficult for the model to focus on hard samples.

To address this problem, we propose a difficulty-aware weight adjustment strategy, which dynamically adjusts the weight of each sample based on its difficulty, as shown in \Cref{fig:method} (b). Specifically, we introduce a difficulty coefficient $\phi  \propto -\textit{mIoU}$ to quantify the difficulty level of each sample, where the function $\phi$ is negatively correlated with \textit{mIoU}. This coefficient dynamically adjusts the sample weights by computing the average accuracy reward of different responses for each sample. The detailed formula is provided below.
\begin{equation}
    \mathcal{J}_{GRPO}(\theta) = \mathbb{E}_{q \sim P(Q), \{o_i\}_{i=1}^G \sim \pi_{\theta_{old}}(O|q)} \left[
    \frac{1}{G}\sum_{i=1}^G \textcolor{brighterblue}{\phi(\mathit{mIoU})} \frac{\pi_{\theta}(o_i|q)}{\pi_{\theta_{old}}(o_i|q)}A_i - \beta\mathbb{D}_{KL}(\pi_{\theta}||\pi_{ref})
    \right]
\end{equation}
For the function $\phi$, we explore various options. This strategy allows the model to pay more attention to difficult samples by assigning more weights to them during the GRPO training, thereby further enhancing its performance.
\section{Experiments}
\subsection{Implementation Details}

\textbf{Datasets.} During training, there are two stages. In the first stage, we jointly utilize 76k CoT cold-start samples from the MGrounding-630k dataset (as mentioned in \Cref{sec:cold-start}) and 14k samples from RefCOCO/+/g~\cite{yu2016refcoco,mao2016refcoco}. In the second stage, we further mix 7k samples from MGrounding-630k with 3k samples from RefCOCO. For evaluation, we assess our model on the multi-image grounding benchmark MIG-Bench~\cite{li2025migician} and the RefCOCO/+/g dataset. Besides, the original MIG-bench dataset contains many incorrect annotations (see more details in the supplementary material), and we manually rectify them. The revised MIG-Bench will be released as well. Additionally, we evaluate the model's zero-shot performance on several benchmarks, including LISA-Grounding~\cite{lai2024lisa}, LLMSeg-Grounding~\cite{wang2024llmseg}, ReVOS Grounding~\cite{yan2024visa}, and ReasonVOS Grounding~\cite{bai2024videolisa}. These datasets are originally designed for segmentation tasks, and we manually extract the corresponding bounding boxes. Among them, LISA and LLMSeg are single-image reasoning grounding tasks, while ReasonVOS and ReVOS are video reasoning grounding tasks. For videos, we uniformly sample 6 frames and require the model to perform grounding on one of these frames.

\textbf{Training Details.} We conduct experiments on both Qwen2-VL-2B and Qwen2-VL-7B models. In the first stage, we use a learning rate of 5e-6 and an accumulated total batch size of 24. In the second stage, the learning rate is set to 1e-6 with an accumulated total batch size of 16. The GRPO algorithm is configured with a maximum completion length of 256 tokens and sampled 8 responses per input.

\textbf{Evaluation Metrics.} We adopt the conventional Acc@0.5 metric for visual grounding tasks. This metric considers a prediction correct if the Intersection over Union (IoU) with the ground truth exceeds 0.5. For all models, we utilize the official checkpoints and conduct evaluations under the same evaluation codes.

\input{tables/main_result}
\input{tables/zero_shot}

\subsection{Main Results}
\textbf{Performance on Migician}. In ~\Cref{tab:main_result}, we present the performance comparison of our UniVG-R1 with Qwen2-VL~\cite{wang2024qwen2vl}, Mantis~\cite{jiang2024mantis}, LLaVA-OV~\cite{li2024llavaov}, MiniCPM2.6~\cite{yao2024minicpm}, mPLUG-Owl3~\cite{ye2024mplug}, InternVL2~\cite{cai2024internlm2} and Migician~\cite{li2025migician} on the MIG-Bench. Our approach achieves new state-of-the-art results across all 10 subtasks, surpassing the previous leading model, Migician, by a significant margin of 9.1\%. Regarding the dataset size, Migician utilizes a total of 1.2 million samples, including 630k from the multi-image grounding dataset, 130k from the RefCOCO dataset, and additional multimodal instruction-following data. In contrast, we only use a curated dataset of 90k CoT samples for stage 1 and 10k for stage 2, totaling 100k samples—approximately 8.3\% of Migician's dataset size. Furthermore, our model significantly outperforms Qwen2-VL-72B by 75.12\%, despite having a much smaller parameter size.

\textbf{Zero-shot performance on reasoning grounding benchmarks}. 
Moreover, ~\Cref{tab:zero_shot} highlights our model's robust zero-shot capabilities. UniVG-R1 consistently achieves superior results across all evaluated reasoning-guided grounding benchmarks, averaging 58.61\% performance on both image and video tasks. While Migician demonstrates stronger performance than Qwen2-VL on video datasets, it lags behind on single-image datasets, particularly Lisa-Grounding. Overall, our model consistently delivers outstanding results on tasks requiring reasoning-chain guidance, excelling in both single-image and multi-image scenarios.

\textbf{Performance on RefCOCO.}
We also evaluate our model on the RefCOCO dataset, as shown in ~\Cref{tab:refcoco}. We compare our UniVG-R1 with VisionLLM v2~\cite{wu2024visionllm}, Shikra~\cite{chen2023shikra}, InternVL2-8B~\cite{cai2024internlm2}, GroundingGPT~\cite{li2024groundinggpt}, Griffon v2~\cite{zhan2024griffon}, GroundingDINO-L~\cite{liu2024groundingdino}, Qwen2-VL-7B~\cite{wang2024qwen2vl}, and Migician~\cite{li2025migician}.
\input{tables/refcoco}
Our model achieves the best average performance of 88.20\%. Notably, we outperform other models on RefCOCOg, which contains more complex reference instructions. This further validates our model's capability to comprehend intricate instructions.

\subsection{Ablation Study}\label{sec:ablation}

\textbf{Training Stages.} Inspired by DeepSeek-R1-Zero, we initially investigate the feasibility of training the model purely through reinforcement learning. As shown in Table~\ref{tab:ablation_stage}, when training on 21k data samples, Pure RL (No.~2) underperforms CoT-SFT (No.~3) by 7.07\% in average score. We attribute this discrepancy to the model's inherent limitations in addressing grounding tasks within multi-image contexts, which makes exploring the reasoning space solely via reinforcement learning particularly challenging. Therefore, we adopt a two-stage training approach.

\textbf{Stage 1:} In this stage, we first examine the effect of data scaling. Increasing the CoT-SFT training dataset from 21k samples (No.~3) to 90k samples (No.~5) improves the average performance by 4.5\%. We also compare standard SFT trained solely with coordinate annotations (No.~4) with CoT-SFT (No.~5), with both models trained on 90k samples. CoT-SFT achieves a higher average performance (69.00\%) compared to SFT (67.30\%). This advantage is particularly evident in the ``Reason'' and ``Co-Re" subtasks, which require strong reasoning abilities. Specifically, CoT-SFT surpasses SFT by 11.34\% in ``Reason'' and 10.27\% in ``Co-Re''. This validates that the reasoning-guided approach enhances the model's reasoning capabilities. After stage 1, CoT-SFT training endows the model with reasoning cognitive abilities.

\textbf{Stage 2:} For Stage 2, all methods are fine-tuned on an additional 10k data samples based on the Stage 1 CoT-SFT model. We compare the performance of continued training with CoT-SFT (No.~6) against employing the GRPO algorithm (No.~7). GRPO improves the average performance by 1.88\% over CoT-SFT. This gain is attributed to GRPO's mechanism of generating multiple responses and assigning different rewards, which guides the model to select the correct reasoning path and thus enhances its reasoning ability. Finally, we compare the standard GRPO algorithm (No.~7) with GRPO eqiupped with our difficulty-aware weight adjustment strategy, referred to as GRPO-Difficulty (No.~8). We observe that this strategy further yields approximately 2.0\% improvement over the standard GRPO, demonstrating the effectiveness of the proposed method.



\textbf{Different difficulty functions.}
Regarding the difficulty-aware weight adjustment strategy proposed in \Cref{sec:difficulty}, we investigate various formulations of the function $\phi$ to modulate sample difficulty. Specifically, we experiment with three distinct functions: $-\log(\textit{mIoU})$, $(1.0 - \textit{mIoU})^2$, and $\exp^{(1 - \textit{mIoU})}$. As shown in ~\Cref{tab:ablation_funtion}, among these, $\exp^{(1 - \textit{mIoU})}$ yields the highest average performance of 72.62\%. Therefore, we adopt this setting as the default in this work.

\input{tables/ablation_stage}
\input{tables/ablation_funtion}
\input{tables/ablation_size}

\textbf{Model size.} 
We also investigate the impact of different model sizes in \Cref{tab:ablation_size}, presenting the performance of Qwen2-VL-2B. Although the 2B model ultimately underperforms compared to the 7B model, GRPO training significantly boosts its performance. We attribute this to the fact that the smaller 2B model may not fully develop its logical reasoning abilities after stage 1 training. As a result, the GRPO algorithm, by guiding the model to select correct reasoning chains, brings about a more substantial performance improvement for the smaller model. Our difficulty-aware weight adjustment strategy further amplifies this gain.

\section{Visualization}
\begin{figure}[t]
    \centering
    \includegraphics[width=\linewidth]{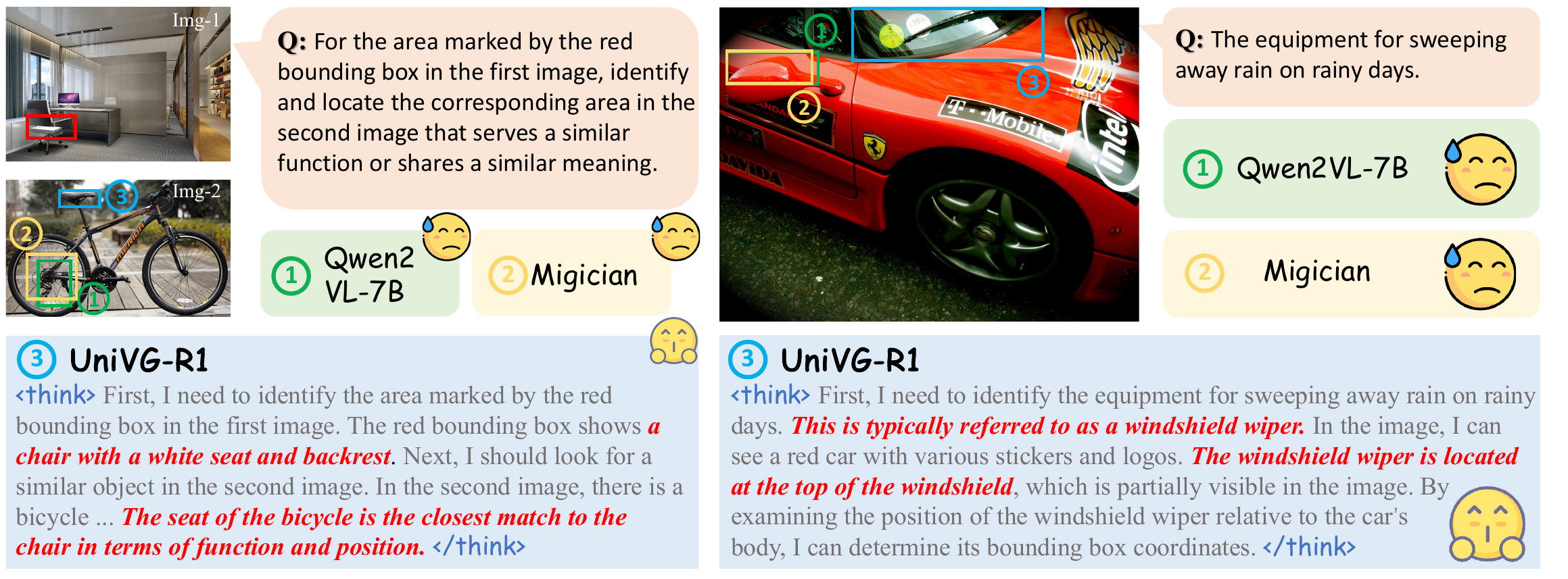}
    \vspace{-12pt}
    \caption{Qualitative comparison of reasoning-guided grounding among Qwen2-VL-7B, Magician, and our UniVG-R1. Left: MIG-Bench. Right: LISA-Grounding.}
    \label{fig:visual}
\end{figure}

In \Cref{fig:visual}, we present a qualitative comparison of our method with Qwen2-VL-7B and Migician. It is evident that UniVG-R1 effectively understands multi-context information across multiple images, as well as implicit instructions (e.g., identifying objects with similar functionality in the left image) and complex instructions (e.g., determining what can sweep away rain in the right image). Compared to other methods, UniVG-R1 provides more accurate results with explanations, demonstrating that our reasoning-guided approach enables the model to better comprehend and execute complex instructions.

\section{Conclusion}
In this work, we propose UniVG-R1, a reasoning-guided MLLM designed for universal visual grounding tasks. UniVG-R1 effectively handles complex textual instructions across diverse multi-modal contexts. To achieve this, we introduce a two-stage training framework: (1) a cold-start supervised fine-tuning stage leveraging a high-quality CoT dataset to guide the model in learning structured reasoning trajectories, and (2) a reinforcement learning stage using the GRPO algorithm to further enhance the model's reasoning capabilities. Furthermore, we propose a difficulty-aware weight adjustment strategy to address the difficulty bias in GRPO training, dynamically prioritizing harder samples to improve overall performance.
Extensive experiments validate the effectiveness of UniVG-R1, which achieves state-of-the-art performance on the multi-image grounding benchmark MIG-Bench with a 9.1\% improvement. Moreover, UniVG-R1 demonstrates strong generalization ability, attaining substantial zero-shot performance gains across multiple reasoning-guided grounding benchmarks. These results highlight the versatility and robustness of our UniVG-R1 in tackling complex, reasoning-guided multimodal grounding tasks.



\bibliographystyle{plain}
\bibliography{ref}


\end{document}

%% file: tables/main_result.tex
\renewcommand{\arraystretch}{1.1}
\begin{table*}[t] 
\centering

\resizebox{\textwidth}{!}{
    \begin{tabular}{l|ccc | ccccccc|c}
    \toprule
    \multirow{4}{*}{\textbf{Models}} & \multicolumn{3}{c}{\textbf{Spontaneous Grounding }} & \multicolumn{7}{|c|}{\textbf{ Referential Grounding }} & \multirow{4}{*}{\textbf{AVG}} \\ 
    \cmidrule(lr){2-4} \cmidrule(lr){5-11} 
    
    \multicolumn{1}{c}{} & \multicolumn{2}{|c|}{\textbf{Difference}} & \textbf{Similarity} &  \multicolumn{4}{c|}{\textbf{Visual Reference}} & \textbf{Textual} & \multicolumn{2}{|c|}{\textbf{Visual+Textual}}& \\ 
    \cmidrule(lr){2-3} \cmidrule(lr){4-4} \cmidrule(lr){5-8} \cmidrule(lr){9-9} \cmidrule(lr){10-11}
    
    \multicolumn{1}{c|}{} & \textbf{Static} & \textbf{Robust} & \multicolumn{1}{|c|}{\textbf{Common}}& \textbf{OT} & \textbf{MV} & \textbf{Region} & \textbf{Refer} &  \multicolumn{1}{|c|}{\textbf{GG}} & \textbf{Reason} & \multicolumn{1}{c|}{\textbf{Co-Re}} & \\
    \midrule
    Qwen2-VL-72B~\cite{wang2024qwen2vl} & 51.13 & 43.61 & 73.74 & 24.54 & 32.63  & 19.86 & 37.37 & 67.83 & 50.51 & 17.94 & 41.91 \\
    \hline
    Mantis~\cite{jiang2024mantis} & 1.52 & 0.00 & 3.31 & 12.18 & 2.08 & 1.00 & 1.01 & 10.02 & 0.00 & 0.85 & 3.20 \\
    LLaVA-OV-7B~\cite{li2024llavaov} & 6.06 & 3.19 & 3.43 & 0.18 & 1.04 & 1.08 & 9.09 & 15.43 & 6.93 & 0.85 & 4.73 \\
    Minicpm2.6~\cite{yao2024minicpm} & 14.58 & 2.13 & 14.34 & 9.82 & 6.25 & 1.75 & 11.11 & 10.02 & 2.97 & 2.56 & 7.55 \\
    mPLUG-Owl3~\cite{ye2024mplug} & 18.56 & 6.38 & 34.93 & 8.55 & 7.64 & 2.41 & 7.07 & 22.85 & 9.09 & 5.98 & 12.35 \\
    InternVL2-8B~\cite{cai2024internlm2} & 8.52 & 19.15 & 38.40 & 19.82 & 10.07 & 5.24 & 34.34 & 39.79 & 26.80 & 7.69 & 20.98 \\
    Qwen2-VL-7B~\cite{wang2024qwen2vl} & 29.92 & 36.17 & 43.07 & 14.55 & 9.38 & 15.54 & 29.29 & 63.51 & 44.33 & 16.24 & 30.20 \\
    Migician~\cite{li2025migician} & 70.64 & 45.74 & 72.76 & 67.82 & 60.07 & 72.57 & 75.76 & 84.12 & 52.58 & 33.33 & 63.54 \\
    \midrule
    \rowcolor{gray!10}
    \textbf{UniVG-R1} & \textbf{71.97} & \textbf{58.51} & \textbf{93.13} & \textbf{76.36} & \textbf{66.32} & \textbf{81.71} & \textbf{82.83} & \textbf{88.04} & \textbf{62.89} & \textbf{44.44} & \textbf{72.64} \\
    \bottomrule
    \end{tabular}
}
\vspace{-3pt}
\caption{Performance comparison on the revised MIG-Bench~\cite{li2025migician}. OT, MV, GG and Co-Re respectively means object tracking, multi-view grounding, group grounding and correspondence. Our UniVG-R1 achieves the best results across all tasks. The best results are shown in bold.}
\label{tab:main_result}
\end{table*}

%% file: tables/zero_shot.tex

\renewcommand{\arraystretch}{0.95}
\begin{table*}[t]
    \centering
    \resizebox{0.95\textwidth}{!}{
    \begin{tabular}{c|ccc|cc|c}
    \toprule
    \multirow{2}{*}{\textbf{Models}} & \multicolumn{3}{c|}{\textbf{Single image}} & \multicolumn{2}{c|}{\textbf{Multi images (video)}} &  \multirow{2}{*}{\textbf{AVG}}\\
    \cmidrule(lr){2-4} \cmidrule(lr){5-6}
     & LISA(val)~\cite{lai2024lisa} & LISA(test)~\cite{lai2024lisa} & LLMSeg~\cite{wang2024llmseg} & ReasonVOS~\cite{bai2024videolisa} & ReVOS~\cite{yan2024visa} &  \\\midrule
    Qwen2-VL-7B~\cite{wang2024qwen2vl} & 52.00 & 49.17 & 35.53 & 9.83 & 23.55 & 34.02 \\
    Migician~\cite{li2025migician} & 36.00 & 32.09 & 34.68 & 33.41 & 39.70 & 35.18 \\
    \rowcolor{gray!10}
    \textbf{UniVG-R1} & \bf 64.00 & \bf 59.69 & \bf 50.60 & \bf 58.73 & \bf 60.03  & \bf 58.61 \\
    \bottomrule
    \end{tabular}
    }
    \caption{Zero-shot performance on several reasoning grounding benchmarks.}
    \vspace{-10pt}
\label{tab:zero_shot}
\end{table*}

%% file: tables/refcoco.tex


\begin{wraptable}{r}{8.5cm}
\centering
\vspace{-16pt}
\resizebox{8.5cm}{!}{

\begin{tabular}{c|ccc|ccc|cc|c}
\toprule
\multicolumn{1}{c|}{\multirow{2}{*}{\textbf{Models}}} & 
\multicolumn{3}{c|}{\textbf{RefCOCO}} & 
\multicolumn{3}{c|}{\textbf{RefCOCO+}} & 
\multicolumn{2}{c|}{\textbf{RefCOCOg}} & 
\multirow{2}{*}{\textbf{AVG}} \\ 
\cmidrule{2-9}
& val & testA & testB & val & testA & testB & val & test & \\
\midrule
VisionLLM v2~\cite{wu2024visionllm} & 79.20 & 82.30 & 77.00 & 68.90 & 75.80 & 61.80 & 73.30 & 74.80 & 74.14\\
Shikra~\cite{chen2023shikra} & 87.00 & 90.60 & 80.20 & 81.60 & 87.40 & 72.10 & 82.30 & 82.20  & 82.97\\
InternVL2-8B~\cite{cai2024internlm2} & 87.10 & 91.10 & 80.70 & 79.80 & 87.90 & 71.40 & 82.70 & 82.70 & 82.94\\ 
GroundingGPT~\cite{li2024groundinggpt} & 88.02 & 91.55 & 82.47 & 81.61 & 87.18 & 73.18 & 81.67 & 81.99 & 83.57\\
Griffon v2~\cite{zhan2024griffon} &89.6& 91.80 & 86.50 & 81.90 & 85.50 & 76.20 & 85.00 & 86.00 & 85.30\\
GroundingDINO-L~\cite{liu2024groundingdino} & 90.60 & 93.20 & 88.20 & 82.80 & 89.00 & 75.90 & 86.10 & 87.00 & 86.60\\
Qwen2-VL-7B~\cite{wang2024qwen2vl} & \textbf{91.70} &  \textbf{93.60} &  \textbf{87.30} &  85.80 &  90.50 &  79.50 &  87.30 &  87.80 &  87.96\\
Migician~\cite{li2025migician} & 91.62 & 93.49 & 87.22& \textbf{86.13} & \textbf{91.06} & 79.93 & 88.06 & 87.80 & 88.16\\ 
\midrule
\rowcolor{gray!10}
\textbf{UniVG-R1} & 91.64 & 93.11 & 87.16 & 85.91 & 90.53 & \textbf{80.04} & \textbf{88.67} & \textbf{88.56} & \textbf{88.20}\\ 
\bottomrule
\end{tabular}
}
\vspace{-8pt}
\caption{The performance on Refcoco/+/g.}
\label{tab:refcoco}
    \vspace{-13pt}
\end{wraptable}

%% file: tables/ablation_stage.tex
\begin{table*}[t] 
\centering

\resizebox{\textwidth}{!}{
\begin{tabular}{c|l|c|ccc | ccccccc|c} 
\toprule
\multirow{4}{*}{\textbf{No.}} & \multirow{4}{*}{\textbf{Methods}} & \multirow{4}{*}{\textbf{\makecell{Data \\ Size}}}  & \multicolumn{3}{c}{\textbf{Spontaneous Grounding }} & \multicolumn{7}{|c|}{\textbf{ Referential Grounding }} & \multirow{4}{*}{\textbf{AVG}} \\
\cmidrule(lr){4-6} \cmidrule(lr){7-13} 

& \multicolumn{1}{c}{} & & \multicolumn{2}{c|}{\textbf{Difference}} & \textbf{Similarity} &  \multicolumn{4}{c|}{\textbf{Visual Reference}} & \textbf{Textual} & \multicolumn{2}{|c|}{\textbf{Visual+Textual}}& \\ 
\cmidrule(lr){4-5} \cmidrule(lr){6-6} \cmidrule(lr){7-10} \cmidrule(lr){11-11} \cmidrule(lr){12-13} 

& \multicolumn{1}{c|}{} & & \textbf{Static} & \textbf{Robust} & \multicolumn{1}{|c|}{\textbf{Common}}& \textbf{OT} & \textbf{MV} & \textbf{Region} & \textbf{Refer} &  \multicolumn{1}{|c|}{\textbf{GG}} & \textbf{Reason} & \multicolumn{1}{c|}{\textbf{Co-Re}} & \\
\midrule
1 & Qwen2-VL-7B & / & 29.92 & 36.17 & 43.07 & 14.55 & 9.38 & 15.54 & 29.29 & 63.51 & 44.33 & 16.24 & 30.20 \\
\midrule

\rowcolor{gray!30}
\multicolumn{14}{c}{\textbf{stage 1}}\\ 
\midrule
2 & Pure RL & 21k & 46.02 & 59.47 & 88.59 & 57.09 & 48.96 & 26.77 & 78.79 & 84.95 & 54.64 & 29.06 & 57.43 \\
3 & CoT-SFT & 21k & 57.58 & 48.94 & 90.18 & 68.91 & 58.68 & 52.78 & 80.81 & 84.54 & 64.95 & 37.61 & 64.50 \\
\midrule 
4 & SFT & 90k & \textbf{73.48} & 45.74 & 89.69 & 71.27 & 62.85 & 86.62 & 77.78 & 84.74 & 49.48 & 31.62 & 67.30 \\
5 & CoT-SFT & 90k & 68.75 & 48.94 & 90.55 & 74.55 & 61.46 & 80.88 & 78.79 & 83.30 & 60.82 & 41.89 & 69.00 \\
\midrule 
\rowcolor{gray!30}
\multicolumn{14}{c}{\textbf{stage 2}}\\ 
\midrule
6 & CoT-SFT & 10k & 70.64 & 52.13 & 89.69 & 75.45 & 60.42 & 76.64 & 78.79 & 83.30 & 58.76 & 41.88 & 68.77 \\
7 & GRPO & 10k & 71.59 & 53.19 & 93.01 & \textbf{77.09} & 64.24 & 80.96 & 81.82 & 86.19 & 55.67 & 42.74 & 70.65 \\
8 & \textbf{GRPO-Difficulty} & 10k & 71.97 & \textbf{58.51} & \textbf{93.13} & 76.36 & \textbf{66.32} & \textbf{81.71} & \textbf{82.83} & \textbf{88.04} & \textbf{62.89} & \textbf{44.44} & \textbf{72.64} \\
\bottomrule
\end{tabular}
}
\vspace{-3pt}
\caption{Ablation study of different stages. We finally adopt CoT-SFT in stage 1, and GRPO equipped with difficulty-aware weight adjustment strategy in stage 2.}
\label{tab:ablation_stage}
\end{table*}

%% file: tables/ablation_funtion.tex
\begin{table*}[t]
\centering

\resizebox{\textwidth}{!}{
    \begin{tabular}{l|c|ccc | ccccccc|c}
    \toprule
    \multirow{4}{*}{\textbf{Methods}} & \multirow{4}{*}{\textbf{\makecell{Difficulty \\ Function}}}  & \multicolumn{3}{c}{\textbf{Spontaneous Grounding }} & \multicolumn{7}{|c|}{\textbf{ Referential Grounding }} & \multirow{4}{*}{\textbf{AVG}} \\ 
    \cmidrule(lr){3-5} \cmidrule(lr){6-12} 
    
    \multicolumn{1}{c}{} & & \multicolumn{2}{c|}{\textbf{Difference}} & \textbf{Similarity} &  \multicolumn{4}{c|}{\textbf{Visual Reference}} & \textbf{Textual} & \multicolumn{2}{|c|}{\textbf{Visual+Textual}}& \\ 
    \cmidrule(lr){3-4} \cmidrule(lr){5-5} \cmidrule(lr){6-9} \cmidrule(lr){10-10} \cmidrule(lr){11-12}
    
    \multicolumn{1}{c|}{} & & \textbf{Static} & \textbf{Robust} & \multicolumn{1}{|c|}{\textbf{Common}}& \textbf{OT} & \textbf{MV} & \textbf{Region} & \textbf{Refer} &  \multicolumn{1}{|c|}{\textbf{GG}} & \textbf{Reason} & \multicolumn{1}{c|}{\textbf{Co-Re}} & \\
    \midrule
    GRPO & / & 71.59 & 53.19 & 93.01 & \textbf{77.09} & 64.24 & 80.96 & 81.82 & 86.19 & 55.67 & 42.74 & 70.65 \\
    \midrule
    GRPO-Difficulty & $-\log(\textit{mIoU})$ & \textbf{72.92} & 54.26 & 92.64 & \textbf{76.91} & 64.93 & 81.55 & \textbf{83.84} & 87.01 & 59.79 & 41.88 & 71.57 \\
    GRPO-Difficulty & $(1.0 - \textit{mIoU})^2$ & 72.73 & 53.19 & 93.12 & 76.36 & \textbf{67.71} & 81.55 & 81.82 & 85.57 & 60.82 & 42.74 & 71.56 \\
    GRPO-Difficulty & $\exp^{(1 - \textit{mIoU})}$ & 71.97 & \textbf{58.51} & \textbf{93.13} & 76.36 & 66.32 & \textbf{81.71} & 82.83 & \textbf{88.04} & \textbf{62.89} & \textbf{44.44} & \textbf{72.64} \\
    \bottomrule
    \end{tabular}
}
\vspace{-3pt}
\caption{Ablation study of funtion $\phi$.}
\label{tab:ablation_funtion}
\end{table*}

%% file: tables/ablation_size.tex
\begin{table*}[t] 
\centering

\resizebox{\textwidth}{!}{
    \begin{tabular}{l|ccc | ccccccc|c}
    \toprule
    \multirow{4}{*}{\textbf{Methods}} & \multicolumn{3}{c}{\textbf{Spontaneous Grounding }} & \multicolumn{7}{|c|}{\textbf{ Referential Grounding }} & \multirow{4}{*}{\textbf{AVG}} \\ 
    \cmidrule(lr){2-4} \cmidrule(lr){5-11} 
    
    \multicolumn{1}{c}{} & \multicolumn{2}{|c|}{\textbf{Difference}} & \textbf{Similarity} &  \multicolumn{4}{c|}{\textbf{Visual Reference}} & \textbf{Textual} & \multicolumn{2}{|c|}{\textbf{Visual+Textual}}& \\ 
    \cmidrule(lr){2-3} \cmidrule(lr){4-4} \cmidrule(lr){5-8} \cmidrule(lr){9-9} \cmidrule(lr){10-11}
    
    \multicolumn{1}{c|}{} & \textbf{Static} & \textbf{Robust} & \multicolumn{1}{|c|}{\textbf{Common}}& \textbf{OT} & \textbf{MV} & \textbf{Region} & \textbf{Refer} &  \multicolumn{1}{|c|}{\textbf{GG}} & \textbf{Reason} & \multicolumn{1}{c|}{\textbf{Co-Re}} & \\
    \midrule
    \rowcolor{gray!30}
    \multicolumn{12}{c}{\textbf{2B model}}\\
    \midrule
    Qwen2-VL-2B & 15.34 & 17.02 & 27.98 & 13.45 & 7.29 & 7.32 & 19.19 & 57.53 & 6.19 & 14.53 & 18.58\\ \midrule
    \ding{172}COT-SFT (90k) & 29.36 & 31.91 & 62.09 & 34.55 & 23.96 & 38.40 & 65.66 & 74.23 & 26.80 & 23.93 & 41.09\\\midrule
    \ding{173}GRPO (10k) & 47.92 & 46.81 & 85.64 & 57.09 & 44.10 & 59.43 & 75.76 & 81.03 & 42.23 & 25.64 & 56.57\\
    \ding{174}GRPO-Difficulty (10k) & 50.57 & 43.62 & 88.34 & 57.09 & 50.69 & 60.27 & 72.73 & 82.06 & 44.33 & 29.91 & 57.96\\
    \bottomrule
    \end{tabular}
}
\vspace{-3pt}
\caption{Performance on 2B model.}
\label{tab:ablation_size}
\end{table*}